\documentclass[letterpaper]{article} 
\usepackage{aaai25}  
\usepackage{times}  
\usepackage{helvet}  
\usepackage{courier}  
\usepackage[hyphens]{url}  
\usepackage{graphicx} 
\usepackage{natbib}  
\usepackage{caption} 
\usepackage{xspace}
\usepackage{comment}
\usepackage{booktabs}       
\usepackage{amsfonts}       
\usepackage{nicefrac}       
\usepackage{microtype}      
\usepackage{subcaption}
\usepackage{siunitx}
\usepackage{amsmath}
\usepackage[ruled,vlined,linesnumbered]{algorithm2e}
\usepackage[inline]{enumitem}
\usepackage[noabbrev]{cleveref}
\DeclareRobustCommand{\abbrevcrefs}{%
\Crefname{appendix}{App.}{Apps.}%
\Crefname{section}{Sec.}{Secs.}%
\Crefname{equation}{Eq.}{Eqs.}%
\Crefname{figure}{Fig.}{Figs.}%
\Crefname{algorithm}{Alg.}{Algs.}%
\Crefname{tabular}{Tab.}{Tabs.}%
\Crefname{lemma}{Lem.}{Lems.}%
\Crefname{corollary}{Cor.}{Cors.}%
\Crefname{theorem}{Thm.}{Thms.}%
\Crefname{proposition}{Prop.}{Props.}%
\Crefname{line}{L.}{Ls.}%
\crefname{appendix}{app.}{apps.}%
\crefname{section}{sec.}{secs.}%
\crefname{equation}{eq.}{eqs.}%
\crefname{figure}{fig.}{figs.}%
\crefname{algorithm}{alg.}{algs.}%
\crefname{tabular}{tab.}{tabs.}%
\crefname{lemma}{lem.}{lems.}%
\crefname{corollary}{cor.}{cors.}%
\crefname{theorem}{thm.}{thms.}%
\crefname{proposition}{prop.}{props.}%
\crefname{line}{l.}{ls.}%
}
\DeclareRobustCommand{\cshref}[1]{{\abbrevcrefs\cref{#1}}}
\DeclareRobustCommand{\Cshref}[1]{{\abbrevcrefs\Cref{#1}}}

\SetKwProg{Fct}{Fct}{}{}

\DeclareMathOperator*{\argmax}{arg\,max}

\newtheorem{definition}{Definition}

\def\cS{{\cal S}}
\def\cA{{\cal A}}

\def\cZ{{\cal Z}}

\def\ie{{\em i.e.}\xspace}

\def\fsc{\mathit{fsc}} 
 
\def\mdp{\textsc{mdp}} 

\newcommand{\eqdef}     {\stackrel{{\textrm{\rm\tiny def}}}{=}}

\mathchardef\mhyphen="2D
\usepackage{accents}
\newcommand{\ubar}[1]{\underaccent{\bar}{#1}}

\usepackage{tikz}
\usepackage{xcolor}
\usepackage[inline]{enumitem}
\usepackage{soul}

\newcommand{\olivier}[1]{\textcolor{blue}{\em [ob] #1}}

\def\depth{\delta}

\title{Partially Observable Monte-Carlo Graph Search}
\author{
    Yang You\textsuperscript{\rm 1},
    Vincent Thomas\textsuperscript{\rm 3},
    Alex Schutz\textsuperscript{\rm 2},
    Robert Skilton\textsuperscript{\rm 1},
    Nick Hawes\textsuperscript{\rm 2},
    Olivier Buffet\textsuperscript{\rm 3}
}
\affiliations{
    \textsuperscript{\rm 1} United Kingdom Atomic Energy Authority\\
    \textsuperscript{\rm 2} Oxford Robotics Institute, University of Oxford\\
    \textsuperscript{\rm 3} Université de Lorraine, CNRS, Inria, LORIA, F-54000 Nancy, France\\

}

\usepackage{bibentry}

\begin{document}

\maketitle

\begin{abstract}
Currently, large partially observable Markov decision processes (POMDPs) are often solved by sampling-based online methods which interleave planning and execution phases.
However, a pre-computed offline policy is more desirable in POMDP applications with time or energy constraints.
But previous offline algorithms are not able to scale up to large POMDPs.
In this article, we propose a new sampling-based algorithm, the \textit{partially observable Monte-Carlo graph search} (POMCGS) to solve large POMDPs offline.
Different from many online POMDP methods, which progressively develop a tree while performing (Monte-Carlo) simulations,
POMCGS folds this search tree on the fly to construct a policy graph,
so that computations can be drastically reduced, and users can analyze and validate the policy prior to embedding and executing it.
Moreover, POMCGS, together with action progressive widening and observation clustering methods provided in this article, is able to address certain continuous POMDPs. 
Through experiments, we demonstrate that POMCGS can generate policies on the most challenging POMDPs, which cannot be computed by previous offline algorithms,
and these policies' values are competitive compared with the state-of-the-art online POMDP algorithms.
\end{abstract}

%

\section{Introduction}

The framework of partially observable Markov decision processes (POMDPs) \citep{ASTROM1965174} provides a mathematical tool for modeling sequential decision-making problems in stochastic and partially observable environments.
%
In a POMDP, the agent needs to take decisions based on its action-observation history or, equivalently, a distribution over possible states (known as a belief).
The objective is to compute a policy that maximizes a given performance criterion. 
%
%
%
%
%
  %
%
However, solving a POMDP is difficult due to the limited information available about the true state.
It has been proven that solving a finite-horizon POMDP is PSPACE-complete \citep{papadimitriou1987complexity}, and finding an optimal policy is undecidable for infinite-horizon POMDPs \citep{madani1999undecidability}.

%
%
%
%
%
%

The majority of state-of-the-art POMDP planners are sampling-based online algorithms \citep{NIPS2010_edfbe1af,NIPS2013_c2aee861} that perform Monte-Carlo simulations in a search tree to plan the best action given the current belief.
The agent then executes this action and updates its belief according to the new observation received before proceeding to the next planning phase.
This interleaves the planning and execution processes, which saves computation resources significantly and enables sampling-based online approaches to address large, and even continuous POMDP problems \citep{Sunberg_2017,Garg2019DESPOTAlphaOP,Hoerger_2021,NEURIPS2021_ef41d488}. 
%
%
However, in applications with constraints such as limited computational power, time, or energy resources, it may not be feasible to perform extensive online computations \citep{Grzes2015EnergyEE}. 
In these situations, a pre-computed policy is often preferable, as it eliminates the need for further computations during execution and can provide immediate responses.
However, offline planning methods encounter scalability challenges due to the computational and memory requirements needed to solve the entire POMDP problem upfront.
This limits the practical use of offline planning for larger or more complex problems, where solving the problem entirely in advance may be infeasible.
%

In this article, taking advantage of successful ideas from existing methods, we propose a novel sampling-based offline algorithm called \textit{Partially Observable Monte-Carlo Graph Search} (POMCGS) with following features:
\begin{enumerate*}
    \item POMCGS performs Monte-Carlo simulations to build and update a complete policy that can be executed offline without any further computations;
    and
    \item POMCGS addresses continuous POMDPs using state discretization, action progressive widening, and observation clustering.
\end{enumerate*}
Experiments show that POMCGS outperforms previous offline methods in scalability and is able to derive compact policies on the most challenging POMDPs, with competitive performance even compared to state-of-the-art online planners.
%


This paper is organized as follows: \Cref{sec:related_work} describes related work on solving POMDPs, and background is presented in \Cref{sec:background}.
We present our contribution in \Cref{sec:POMCGS}, and give experiments with comparisons to state-of-the-art POMDP solvers in \Cref{sec:experiments}.
\Cref{sec:discussion} discusses the strengths and limitations of our algorithm, as well as opportunities for related research fields.
Finally, we conclude this work in \Cref{sec:conclusion}.

\section{Related Work}
\label{sec:related_work}

 In this section, we review work related to our contribution.
 %
 \citet{lauri2022partially} and \citet{kurniawati2022partially} present more detailed surveys about POMDPs.
 %
%
%

%
Offline POMDP solvers aim to compute a policy anticipating all possible situations.
In this category, {\em point-based} methods \citep{Pineau-ijcai03, Smith-uai04,Smith_HSVI2,sarsop, mcvi} approximate the optimal value function by exploiting its convexity, and focus on representative belief points that are reachable under an optimal policy;
{\em policy iteration} algorithms \cite{Hansen-nips97,hansen98,Poupart2003,Ji2007PointBasedPI} iteratively refine a policy represented by a finite-state controller (FSC) until convergence in small scale POMDP problems;
while {\em expectation-maximization} methods \citep{toussaint2006probabilistic, NIPS2011_1f3202d8} turn a POMDP into a probabilistic inference problem and estimate its parameters, which only gives local optima.
%

Conversely, state-of-the-art POMDP solvers are usually online methods based on Monte-Carlo Tree Search \citep{UCT}.
In this category, {\em partially observable Monte-Carlo planning} (POMCP) \citep{NIPS2010_edfbe1af} extends MCTS to large POMDPs.
POMCP performs forward simulations in a search tree, and each tree node corresponds to an action-observation history.
Alternatively, {\em determined sparse partially observable tree} (DESPOT) \citep{NIPS2013_c2aee861} uses a fixed number of scenarios to sample observations and maintain a sparse belief-tree, which proves to be very efficient for solving POMDPs with many observations. 
These online algorithms outperform the previous offline methods in terms of scalability.
Recent works \citep{Sunberg_2017,Garg2019DESPOTAlphaOP,Hoerger_2021,NEURIPS2021_ef41d488} extend online POMDP planning further to continuous domains.
For example, \citet{Sunberg_2017} extend POMCP by using double progress widening during the tree search, \ie, slowly expanding the sets of applicable actions and considered observations.
%
%

One can improve the efficiency of MCTS-based methods by exploiting the continuity of the optimal value function in state or belief space \cite{Paludo2024ECAI}.
%
For instance, performance can be improved by merging similar belief nodes at the same depth along the search tree \citep{DBLP:conf/icra/BallesterosMTVO13, NEURIPS2021_ef41d488,kujanpaa2023continuous}, which avoids expanding unnecessary branches.
\citet{pmlr-v129-leurent20a} discuss another variation called Monte-Carlo graph search in fully observable settings that merges nodes corresponding to similar states reached via different trajectories.
An important insight for POMDPs is that merging similar nodes in a search tree leads to a structure similar to a finite state controller (FSC).
%
%
%
%
In this article, we propose a Monte-Carlo graph search algorithm in POMDPs which, by merging similar nodes at possibly different depths, constructs an FSC on the fly.
\section{Background}
\label{sec:background}

\subsection{POMDPs}
%
%
%
%
In this work, we consider one agent interacting with an environment under uncertain dynamics and partial observability.
%
%
Such a problem is described as a {\em partially observable Markov decision process} (POMDP), which is formally defined by a tuple $\langle \cS, \cA,\Omega,T,O,r,b_{0}\rangle$  where
  \begin{enumerate*}[label=(\roman*), nosep]
  \item $\cS$ is the set of states;
  \item $\cA$ is the set of actions;
  \item $\Omega$ is the set of observations;
  \item $T(s,a,s') = Pr(s' | s, a)$ is the transition function, which encodes the dynamics of the environment (the probability of reaching the next state $s'$ given the current state $s$ and an executed action $a$);
  \item $O(o, a, s') = Pr(o | s', a)$ is the observation function; it indicates the probability of receiving an observation $o$ given the next state $s'$ reached while performing action $a$;
  \item $r(s,a)$ is the reward function, which gives the instant reward when performing action $a$ at state $s$;
  \item $b_{0}$ is an initial probability distribution over states.    
  \end{enumerate*}

%
%

In POMDPs, a policy $\pi$ maps action-observation {\em histories} $h_t=\langle a_1, o_1, \dots, a_t, o_t \rangle$ to actions, and its {\em value function} gives the expected discounted cumulative rewards under $\pi$ from any reachable history: $V^\pi(h_t) \eqdef \sum_{k=t}^\infty \mathbb{E}_\pi\left[ \gamma^{t-k} r(S_k,\pi(H_k)) | H_t=h_t \right]$,
and the action-value is defined as $Q^\pi(h,a) \eqdef \mathbb{E}_\pi \left[ r(S_t,A_t) + \gamma V^\pi(H_{t+1}) | H_t=h_t, A_t=a \right]$.
There exists at least one {\em optimal policy} $\pi^*$, \ie, a policy with {\em optimal value function} $V^*(h_t)=\max_\pi V^\pi(h_t)$ for any $h_t$.
The probability distribution over states under history $h_t$ (obtained through Bayesian inference) is its {\em belief state} $b$.

%
%
Explicit transition, observation and reward functions may not be available, or may be impractical due to time or memory requirements.
%
In this case, one may rely on a generative model (black box simulator) $\mathcal{G}$, which samples a triplet $\langle s', o, r \rangle$ given a state-action pair $\langle s, a \rangle$.
\subsection{POMCP}
{\em Partially observable Monte-Carlo planning} (POMCP) \citep{NIPS2010_edfbe1af} is a sampling-based POMDP solver that extends MCTS to solve large discrete POMDPs.
It maintains a partial search tree online during execution, each node corresponding to an action-observation history.
Starting from the current root node under $h_t$, POMCP samples states from an estimated belief $\hat{b}_t$ and simulates trajectories in the search tree.
Along such a trajectory, it uses a bandit-based strategy to favor actions with high estimated value while still refining estimates of all action values.
At the end of each trajectory, the search tree is locally expanded, and a heuristic is used to estimate the new leaf's value.
When visiting a node, the current state is stored, thus contributing to that node's belief estimate $\hat{b}_t$, which will serve in particular if, after an actual transition, the node becomes the new root node.

\subsection{Finite State Controllers}
\label{sec:FSC_def}

%
%
Sampling-based solvers such as POMCP typically build a search tree, and could thus be used in finite-horizon problems to compute policy trees offline.
%
However, for an infinite-horizon setting, a policy tree will have infinitely many nodes, which cannot be all visited and stored.
In infinite-horizon POMDPs, solution policies are better represented as {\em finite state controllers} (FSC) (also called {\em policy graphs} \citep{MeuKimKaeCas-uai99}),
\ie, automata whose transitions from one internal state to the next depend on the received observations and generate the actions to be performed.
%

\begin{definition}
  For some POMDP sets $\cA$ and $\Omega$, %
  a (deterministic) {\em FSC} is specified by a tuple
  $\fsc \equiv \langle \mathcal{N}, \eta, \psi \rangle$, where:
  \begin{enumerate*}[label=(\roman*), nosep]
    \item $\mathcal{N}$ is a finite set of nodes, %
    with $n_0$ the start node; %
  \item $\eta: \mathcal{N} \times \cA \times \Omega \to \mathcal{N} $ is the node transition function; %
    $n'=\eta(n,\langle a, o \rangle)$ is the next node after executing action $a$ and observing $o$ from node $n$;
  \item $\psi: \mathcal{N} \to \cA $ is the action-selection function of the FSC; %
    $a=\psi(n)$ is the action triggered when in node $n$.
  \end{enumerate*}

\end{definition}

In this article, we represent POMDP policies as FSCs (as described in the next section).

%

\section{Monte-Carlo Graph Search for POMDPs}
\label{sec:POMCGS}
This section presents our algorithmic contribution.
As an offline solver, POMCGS computes a compact policy represented by an FSC.
%
Its main algorithm is given in \Cref{alg:main_algo}.
%
%
%
First, POMCGS
\begin{enumerate*}[label=(\roman*), nosep]
    \item initializes some parameters (\cref{alg|main|inits});
    \item solves the underlying MDP (\ie, assumes full observability of the POMDP problem) to provide an upper-bounding value function $V_\mdp$ (\cref{alg|main|Vmdp}); and
    \item creates an FSC policy $\pi_{\fsc}$ with a start node $n_0$ labeled by the initial belief $b_0$  (\cref{alg|main|create_init_node,{alg:add_init_node},alg|main|createFSC}).
\end{enumerate*}
%
POMCGS then iteratively improves (\cref{alg|main|improve}) and evaluates (\cref{alg|main|evaluate}) its FSC policy until convergence, 
%
%
before pruning unreachable nodes in the solution FSC using a graph traversal from $n_0$ to identify reachable nodes (\cref{alg|main|prune}).
%

%
%

%
%
\Cref{sec:merge_belief} will discuss how to efficiently identify similar beliefs,
before describing how to construct an FSC on the fly in \Cshref{sec:UpdateFSC}.
\Cshref{sec:bounds} then explains POMCGS's stopping criterion,
and \Cshref{sec:continuous_domains} presents the methods we use to address continuous domains.
%

%
\begin{algorithm}
  \caption{POMCGS Main Algorithm}
  \label{alg:main_algo}
  \DontPrintSemicolon
  [Initialization:] \label{alg|main|inits} %
%
Define $\epsilon$ and $\xi$ for convergence detection and belief merging, and
$nb_{eval}$ and $nb_{sim}$ for policy evaluation and improvement. \;

Compute $V_\mdp$ by solving the underlying MDP \label{alg|main|Vmdp}\;

$n_0 \gets node(b_0) $  \label{alg|main|create_init_node} \; 
$\mathcal{N}_{\fsc} \gets \{n_0 \}$  \label{alg:add_init_node} \;
$\pi_{\fsc} \gets \langle \mathcal{N}_{\fsc}, \psi, \eta \rangle$ \label{alg|main|createFSC} \;
\While{$(\bar{V}(b_0)_{\pi_{\fsc}} - \ubar{V}(b_0)_{\pi_{\fsc}}) > \epsilon$}
{
    $\text{UpdateFSC}(\pi_\fsc, b_0, \xi, nb_{sim})$ \label{alg|main|improve}  \;
    $\text{EvaluateFSC}(\pi_\fsc, b_0, nb_{eval})$ \label{alg|main|evaluate} \;
}
$\text{Pruning}(\pi_{\fsc})$ \label{alg|main|prune} \;
\Return{$\pi_{\fsc}$}
\end{algorithm}
%

\subsection{Merging Similar Belief Nodes Efficiently}
\label{sec:merge_belief}


A POMDP's optimal value function $V^*$ is continuous in belief space \cite{SmaSon-or73}.
Consequently, two nodes of the policy tree with similar beliefs can be merged with limited harm to the resulting policy \cite{hsu2007makes}.

%
To enable merging, POMCGS needs each new node to be equipped with a well-estimated belief (using many particles) from the beginning, in contrast to POMCP.
%
When facing a new belief (estimate) $b$,  the norm-1 distance between beliefs is used to search for an existing node $n$ (if one exists) such that
\begin{align}
\label{eq:belief_compare}
    \|b-n.b\|_1 \leq \xi,
\end{align}
with $\xi>0$ a threshold.
%
%
As a naive search would be computationally expensive for large FSCs, nodes are stored in a data structure similar to cover trees \citep{Beygelzimer2006,izbShe-icml15}, which significantly reduces the number of comparisons, and also allows for cheap insertions of new nodes.
%
%
%
This process (called \textit{SearchOrInsert} in \Cref{alg:UpdateFSC}) takes the current set of FSC nodes $\mathcal{N}$ and a new belief $b'$ as input, then returns either:
%
%
\begin{enumerate*}[label=(\roman*), nosep]
\item an existing node $n$ whose norm-1 distance satisfies \Cref{eq:belief_compare}, or
\item a new created node with $b'$ since there is no similar belief found in the FSC. 
\end{enumerate*}
Another function, \textit{SearchNode}$(b',\mathcal{N})$, searches for the node in $\mathcal{N}$ closest to $b'$.


\subsection{Constructing an FSC through Monte-Carlo Simulations}
\label{sec:UpdateFSC}

The policy improvement method \textit{UpdateFSC} is given in \Cshref{alg:UpdateFSC}.
As in POMCP, we relax the need for an exact model by performing Monte-Carlo simulations with a black-box simulator.
But, unlike POMCP, our method directly builds and searches in a graph (finite-state controller) instead of a tree.
This process starts from the initial belief $b_0$ and repeatedly samples states to simulate $nb_{sim}$ trajectories (lines~\ref{alg:POMCP_repeat_start}--\ref{alg:POMCP_repeat_end}), before returning the current FSC in \cref{alg:return_FSC}.
%
%
Trajectories are generated by the recursive \textit{Simulate} function,
which stops when $\frac{\gamma^\depth}{1-\gamma}(r_{\max}-r_{\min}) \leq \epsilon$,
where $\depth$ is the current depth, and the reward is bounded in $[r_{\min},r_{\max}]$.

In \textit{Simulate},
an action $a$ is selected 
using the UCB rule in \cref{alg:defaultUCB}. 
%
%
%
%
%
Visit counts $N(n)$ and $N(n,a)$ of the current node $n$ are updated {\em before} moving on to the next node (lines~\ref{alg:UpdateNodeVisit}--\ref{alg:UpdateNodeActionVisit}).
This is required to avoid repeating the same decision if node $n$ is visited twice along the same trajectory, which may happen in such a graph with loops.
%
%
%
Then, if $a$ is performed for the first time in $n$, a call to \textit{ProcessAction} (detailed in the next paragraph) creates outgoing connections for each observation and returns a first estimate of $a$'s action value.
If $a$ has already been visited in $n$, POMCGS simulates $a$ with the sampled state $s$, derives the next FSC node $n'$, updates $n$'s $Q$-value, and assigns $\psi(n)$ with $\argmax_a(Q(n,a))$ (lines~\ref{alg:simulate_sa}--\ref{alg:end_simulate}) before returning the trajectory's estimated return $R$.
%
%

\textit{ProcessAction} 
simulates $a$ in $n$'s belief $n.b$ repeatedly until enough particles are collected to build the next nodes (lines~\ref{alg:ProcessActionRepeatStart}--\ref{alg:ProcessActionRepeatStop}).
%
%
%
Each encountered observation (gathered in a set in
\cref{alg|ProcessAction|getObservations}) is attached to its belief estimate (map created in \cref{alg|ProcessAction|BuildBeliefs}).
%
Then, for each reachable belief $b'$,
if the FSC has reached its size limit $N_{max\mhyphen fsc}$, we choose the next node $n'$ in $\mathcal{N}$ minimizing 
$\|n'.b - b' \|_1$ (\cref{alg:FindSimiliarNode}).
%
%
%
Otherwise, we search for a node whose belief is $\xi$-close to $b'$ in norm-1 distance, and, if none exists, we create a next node $n'$ with belief $b'$ and add it to the FSC (\cref{alg:CreateNewNode}).
%
If $n'$ is created, then its value is initialized optimistically using a $V_\text{MDP}$ heuristic \citep{Hauskrecht-jair00} (line~\ref{alg:NodeRollout}).
%
%
In line~\ref{alg:InitNodeQ}, node $n$'s $Q$-value is initialized with the estimate of the instant reward, plus the discounted future values.


\begin{algorithm}[t!]
  \caption{FSC Improvement Procedure}
  \label{alg:UpdateFSC}
  \DontPrintSemicolon
\Fct{\text{UpdateFSC}($\pi_\fsc, b_0, \xi, nb_{sim}$)
\label{alg:search_start}}
{
  \For{$i\in 1:nb_{sim}$ \label{alg:POMCP_repeat_start} }
  {
    $s \sim b_0$   \;
    \textit{Simulate}($s, n_0, 0$) \label{alg:simulate} \;
  }\label{alg:POMCP_repeat_end}

    \Return{$\pi_\fsc$} \label{alg:return_FSC}
}


\Fct{Simulate($s, n, \depth$)}
{

  \lIf{ $\frac{\gamma^\depth}{1-\gamma}(r_{\max}-r_{\min}) < \epsilon$}
  {
    \Return{$0$}
  }


    $a \gets \argmax_{a \in C(n)} \left[ Q(n,a) + c\sqrt{\frac{\log N(n)}{N(n,a)}} \right] $ \label{alg:defaultUCB} \; %
  $N(n) \gets N(n) + 1$  \label{alg:UpdateNodeVisit}  \;
  $N(n, a) \gets N(n, a) + 1$ \label{alg:UpdateNodeActionVisit}\;
  \eIf{$N(n, a) = 1$}{
    \Return{ProcessAction($n, a, \depth$)} 
  }{
    $(s', o) \sim \mathcal{G}(s, a)$ \label{alg:simulate_sa} \;
    $n' \gets \eta(n, \langle a, o \rangle)$ \label{alg:get_next_node} \;
    $R \gets r(n, a) + \gamma \textit{Simulate}(s',  n', \depth + 1)$  \;
    $Q(n, a) \gets Q(n, a) + \frac{R - Q(n, a)}{N(n, a)}$ \;
    $\psi(n) \gets \argmax_{a} Q(n,a)$ \label{alg:end_simulate} \;
    \Return{$R$}
  }
}

\Fct{ProcessAction($n, a, \depth$)}
{
  $r(n, a) \gets 0$ \; 
  $\mathcal{P} \gets \emptyset$ \; 
  \Repeat{$ |\mathcal{P}| \geq nb_{\text{particles}} $ }{ \label{alg:ProcessActionRepeatStart}
    $s \sim n.b$ \;
    $\langle s', o, r  \rangle \sim\mathcal{G}(s, a) $ \;
    $r(n, a) \gets r(n, a) + r$ \;
    $ \mathcal{P} \gets \mathcal{P} \cup \langle 
 s', o \rangle$ \;
  } \label{alg:ProcessActionRepeatStop}
    $r(n, a) \gets \frac{r(n, a)}{|\mathcal{P}|}  $ \;
    $\tilde{\Omega} \gets \mathcal{P}.\{o\}$ \label{alg|ProcessAction|getObservations} \;
  $B' \gets \textit{BuildBeliefs}(\mathcal{P}.\{ s'\}, \tilde{\Omega}) $ \label{alg|ProcessAction|BuildBeliefs} \;
  \For{$\tilde{o} \in \tilde{\Omega}$}
  {
    $b' \gets B'[\tilde{o}]$ \;
    \eIf{$|\mathcal{N}| = \mathcal{N}_{max\mhyphen \fsc} $} 
    {
      $n' \gets \textit{SearchNode}(b', \mathcal{N})$  \label{alg:FindSimiliarNode}
    }{
      $n' \gets \textit{SearchOrInsert}(b', \mathcal{N}, \xi)$\label{alg:CreateNewNode}\;
        \If{$N(n')=0$}{
            $V(n') \gets \sum_{s}b'(s)V_\mdp(s)$ 
            \label{alg:NodeRollout} \;
        }
    }
    $\eta(n, \langle a, \tilde{o} \rangle) \gets n'$
  }
  
    $Q(n, a) \gets  r(n, a) + \gamma \sum_{\tilde{o}} \frac{|B'[\tilde{o}]|}{|B'|} V(n')$ \label{alg:InitNodeQ} \;
    \Return{$Q(n,a)$}
}
\end{algorithm}

Note that the solution FSC may be incomplete, \ie, with leaves that have not been visited yet.
If such a leaf is reached at execution time, then the blind policy is executed to guarantee a lower-bound value (see next section).
If a pre-computed FSC is desired to have some understanding of the policy prior to executing it, but computational resources are still available at execution time, then an online solver could be used once a leaf of the FSC is reached (taking over control from then on, using the leaf's attached belief as initial belief), rather than relying on a blind policy to achieve better performance.

\subsection{Estimating Upper and Lower Bounds}
\label{sec:bounds}
We propose regularly estimating upper- and lower-bounds of the current FSC's value at $b_0$ (denoted $\bar{V}$ and $\ubar{V}$), using the function \textit{EvaluateFSC} described in \Cref{alg:Evaluation}, to stop POMCGS when the gap between bounds is small enough, despite the lack of guarantees w.r.t. $V^*(b_0)$.
These estimates are obtained through $nb_{eval}$ Monte-Carlo simulations of the FSC (\cref{alg|fscEval|loop}), stopping trajectories
either when the remaining expected return is below a threshold $\epsilon>0$ (\cref{alg|fscEval|epsilon}), or
upon encountering a node $n$ with an insufficient visit count (\cref{alg|fscEval|visitCount}), \ie, $N(n) \geq N^*$ (with $N^*\geq 1$, so that the simulation stops before leaving the FSC if it is incomplete).
In the latter case, the value at $n$ is
upper-bounded with a $V_\text{MDP}$ heuristic  (\cref{alg|fscEval|Vmdp}),
and
lower-bounded using a blind policy approximation (\cref{alg|fscEval|blind}) \citep{Hauskrecht-jair00}, which repeatedly selects the action that maximizes the worst-case reward.

Theoretically, both estimates asymptotically converge to $V_{fsc}(b_0)$ upon increasing the number of simulations $nb_{eval}$ (and decreasing $\epsilon$) only for FSCs where $N(n)\geq N^*$ for any reachable node $n$, implying that these FSCs are ``closed", \ie, without open nodes.
We use large values of $N^*$ expecting that decisions will be optimal in any node $n$ with $N(n)\geq N^*$.
However this also requires having good enough belief estimates (\ie, large $nb_{\text{particles}}$) and few merges (\ie, small $\xi$).
Under these conditions, POMCGS shall stop with an FSC that is $\epsilon$-optimal in discrete problems.
\begin{algorithm}[t]
  \caption{FSC Evaluation Procedure}
  \label{alg:Evaluation}
  \DontPrintSemicolon
\Fct{$\text{EvaluateFSC}(\pi_\fsc, b_0, nb_{eval})$
\label{alg:estimate_start}}
{
    $\bar{V}, \ubar{V} \gets 0$ \;
    \For{$i \in 1: nb_{eval}$ \label{line:stop_ub_simulation} \label{alg|fscEval|loop}}{
    $s \sim b_0$ \;
    $\delta \gets 0$ \;
    $n \gets n_0$ \;
    \While{$\frac{\gamma^\depth}{1-\gamma}(r_{\max}-r_{\min}) < \epsilon$ \label{alg|fscEval|epsilon}}
    {
        \eIf{$N(n) > N^*$ \label{alg|fscEval|visitCount}}
        {
            $a \gets \psi(n)$ \;
            $(s', o, r) \sim \mathcal{G}(s, a)$ \;
            $n \gets \eta(n, \langle a, o \rangle)$ \;
            $\bar{V} \gets \bar{V} + \gamma^{\delta} r$ \;
            $\ubar{V} \gets \ubar{V} + \gamma^{\delta} r$
        }
        {
            $\bar{V} \gets \bar{V} + \gamma^{\delta} \sum_{s \in n.b} n.b(s) \cdot V_\mdp(s)$ \label{alg|fscEval|Vmdp} \;
            $\ubar{V} \gets \ubar{V} + \gamma^{\delta} \frac{1}{1-\gamma} {\max_a\min_s r(s, a) }$ \label{alg|fscEval|blind} \; 
            break \;
        }      
        $\delta \gets \delta + 1$ \;
    }	
}   

\Return{$\frac{\bar{V}}{nb_{eval}}, \frac{\ubar{V}}{nb_{eval}}$}
}
\end{algorithm}

\subsection{Handling Continuous POMDPs}
\label{sec:continuous_domains}
We now describe how POMCGS is modified to handle continuous POMDPs, where any combination of state, action and observation spaces could be continuous.

Continuous states are an issue only when it comes to measuring the distance between two belief estimates.
In this case, the state space is discretized, so that particles are grouped in bins, which allows still using the same distance.
A dedicated metric for continuous distributions, such as the Wasserstein distance, could allow for better merges, but may be much more time consuming.
%

Continuous actions are handled through action progressive widening (APW) as presented in \Cref{alg:APW}, which is called instead of \cref{alg:defaultUCB} of \Cref{alg:UpdateFSC}.
APW is stopped in nodes considered ``finalized" ($N(n)\geq N^*$), so that the number of actions is bounded.
One could also sample a set of actions initially, but it is preferable to use progressive widening to ensure that a few actions are well estimated rather than obtaining poor estimates for many actions.
%
\begin{algorithm}[t]
\caption{Action Progressive Widening}
  \label{alg:APW}
  \DontPrintSemicolon
\Fct{\text{ActionProgressiveWidening}($n$)}
{
    \eIf{$|C(n)| \leq k_{a} N(n)^{\alpha_a}$ 
 \text{and}  $N(n) < N^*$ }
    {
        $a \sim A $ \;
        $C(n) \gets C(n) \cup a $
    }
    {
         $a \gets \argmax_{a \in C(n)} [ Q(n,a) + c\sqrt{\frac{\log N(n)}{N(n,a)}} ] $ \label{alg:UCB} \;
    }
}   
\Return{$a$}
\end{algorithm}
%
\begin{figure}
  \centering
    \includegraphics[width=0.99\columnwidth]{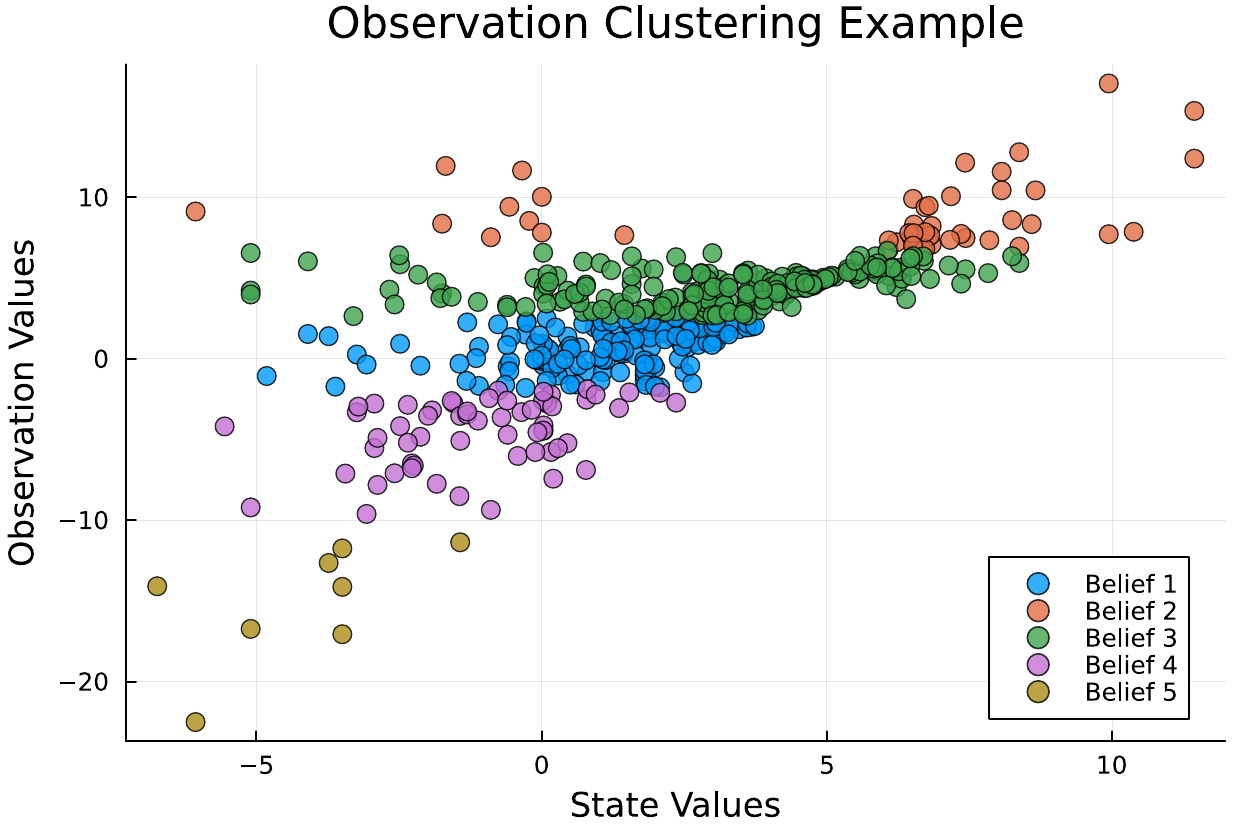}
  \caption{An illustration of observation clustering using \textit{$K$-Means} in the Light Dark Problem \citep{Sunberg_2017}. 
    Each point corresponds to a particle pair of observations (value on the $y$ axis) and new states (value on the $x$-axis) collected by performing an action $a$ from some belief $b$. 
    Particle pairs are clustered according to the observation values into $K=5$ labels represented by different colors, and beliefs are created with states attached to the same label.}
  \label{fig:Kmeans_obs}
\end{figure}

Continuous observation spaces need to be discretized because the FSC needs to anticipate all possible observations.
This is different from online algorithms such as POMCPOW, which can rely on observation progressive widening for exploration and then adapt to actual observations at execution time.
Here, we propose to use \textit{$K$-means} \cite{MacQueen1967} instead of \cref{alg|ProcessAction|getObservations} of \Cref{alg:UpdateFSC} to cluster observations independently in each created node.
When simulating an action $a$ from some belief $b$ in function \textit{ProcessAction} (\Cshref{alg:UpdateFSC}),
observation and new state pairs $(o,s')$ are collected and clustered by observation values according to a pre-defined parameter $K \in \mathbb{N}^*$, 
and new beliefs are created with their corresponding states (\cshref{alg|ProcessAction|BuildBeliefs}).
An example of this clustering is illustrated in Fig.~\ref{fig:Kmeans_obs}.
Note that, unlike POMCPOW, we do not need an explicit observation model to query the relative likelihood of different observations, a black-box simulator suffices.
\section{Experiments on POMDP Benchmarks}
\label{sec:experiments}


In this section, we evaluate POMCGS on various benchmarks provided by the POMDP.jl framework \citep{pomdpjl}, 
%
and compare our methods with other state-of-the-art offline solvers:
SARSOP \citep{sarsop} and MCVI \citep{mcvi};
and online planners:
DESPOT \citep{NIPS2013_c2aee861}, POMCPOW \citep{Sunberg_2017}, and AdaOPS \citep{NEURIPS2021_ef41d488}.
Note that SARSOP is usually considered as the best $\epsilon$-optimal offline POMDP solver and often used to calibrate other online methods in small or moderate-size POMDPs.
However SARSOP requires explicit POMDP models while other solvers do not.

All experiments for POMCGS were conducted on a computer with a 4.8\,GHz i7 \textsc{cpu} and 64\,GB \textsc{ram}.
Source code and parameter settings are provided in the supplementary material.
In this article, $V_\mdp$ is computed by Q-learning on the underlying MDP, which is used to estimate POMCGS' upper bound (for continuous domains, Q-learning requires domain discretization). 
One can also use other MDP techniques to derive $V_\mdp$ \cite{kochenderfer2015decision}, which is a mapping from states to values. 
In POMCGS, we set $nb_{sim} = 10^3$ in \textit{UpdateFSC} for the policy improvement, and $nb_{eval} = 10^5$ in \textit{EvaluateFSC} to evaluate the policy's upper and lower bounds with $N^* = 50$.
For each POMDP benchmark, we ran POMCGS 10 times to compute the average return.
We set $\epsilon = 0.01$ to detect convergence.\footnote{
    The UCB constant $c$ was set
    to $2.0$ for RS(7,8), RS(11, 11), RS(15, 15) and Light-Dark; and
    to $1.0$ for Bumper Roomba, Lidar Roomba and Laser Tag.
    We do not further discuss the influence of this parameter, which appeared not to be significant.
}
%
All online solvers are limited to 3 seconds of computation time per step.

\Cref{table1} presents the average discounted returns with associated standard errors for the 7 benchmark problems evaluated.
%
%
SARSOP is labeled with a $*$ exponent indicating that an explicit POMDP model is required. 
For POMCGS, we report its final average lower bound.
Entries with a ``\text{---}'' symbol indicating that the algorithm is not able to run in this domain.

%
In terms of value achieved, on small and moderate-size domains,
POMCGS is able to achieve near-optimal performance as SARSOP; 
on large and continuous problems (except the Laser Tag domain), POMCGS is the only offline algorithm to compete with state-of-the-art online algorithms.
In particular, although both POMCGS and MCVI are able to derive FSC solutions among all offline solvers, there is no belief merging in MCVI.
In each MCVI backup, it needs to add one node and compute a new FSC entirely, which is computationally expensive. 
Therefore, given a fixed timeout, we observed that MCVI's performance degrades when the required planning horizon increases on large problems. 
In contrast, POMCGS merges similar nodes, which largely reduces unnecessary computations.

%
%

In later sections, we analyze POMCGS's performance individually for each domain and discuss the influence of parameters.

\begin{table*}
  \centering
 \resizebox{1.0\linewidth}{!}{
    \begin{tabular}{lrrrrrrr}
    \toprule
           & {RS$(7, 8)$} &  {RS$(11, 11)$} &  {RS$(15, 15)$} & Light Dark  & Bumper Roomba & Lidar Roomba & Laser Tag  \\
    \midrule
    $|\cS|$ & $12,544$ & $247,808$ & $7,372,800$ & $\infty$ & $\infty$ & $\infty$ & $5,830$\\
    $|\cA|$ & $13$ & $16$ & $20$ & $3$ & $\infty$ & $\infty$ & 5\\
    $|\cZ|$ & $3$ & $3$ & $3$ & $\infty$ & $2$ & $\infty$ & $ \sim1.5 \times 10^6 $ \\   
    \midrule
    \textit{Offline Solvers} \\
    \textbf{POMCGS} & $21.13 \pm 1.09$ & $19.24 \pm 1.31 $ & $16.28 \pm 1.44$ & $3.74 \pm 0.06$ & $0.31 \pm 0.18 $ & $0.79 \pm 0.12$ & $-19.42 \pm 0.15$ \\ 
    MCVI  & $19.36 \pm 1.21$ & $18.09 \pm 1.25$ & $13.70 \pm 0.96$  & $2.94 \pm 0.15$ & \text{---} & \text{---} & \text{---}   \\
    SARSOP* & $21.45 \pm 0.08$ &  $21.56 \pm 0.12$ & \text{---} & \text{---} & \text{---} & \text{---} & \text{---}	\\
    \midrule
    \textit{Online Planners} \\
    AdaOPS & $20.93 \pm 0.35$  & $20.17 \pm 0.41$ & $17.81 \pm 0.43$ & $3.76 \pm 0.10$ & \text{---} & \text{---} & $-8.82 \pm 0.21$  \\
    DESPOT & $21.08 \pm 0.21$ & $20.55 \pm 0.26$ & $18.32 \pm 0.29$ & $2.66 \pm 0.14$ & \text{---} & \text{---} & $-9.14 \pm 0.27$   \\
    POMCPOW & $20.55 \pm 0.17$ & $19.51 \pm 0.19$ & $16.57 \pm 0.23$ & $3.29 \pm 0.08$  & $-0.41 \pm 0.15$ & $0.73 \pm 0.10$ & $-9.46 \pm 0.24$ \\
    \bottomrule
    \end{tabular}
}
  \caption{Performance Comparison of Different POMDP Planning Algorithms}
\label{table1}
\end{table*}

\subsection{Benchmark Problems}

\subsubsection{Rock Sample:}
\citet{Smith-uai04} proposed a large-state benchmark called \textit{RockSample} (RS), in which a robot moves in a grid-world and tries to collect rocks. %
In RS($n, k$), the map is defined as an $n \times n$ grid and contains $k$ rocks.
Each rock may be good or bad while the robot is only equipped a noisy sensor to test it. 
The robot's goal is to collect as many good rocks as possible and exit on the east boundary of the map. 
%

%
%
%
We set offline solvers with timeouts of 1 hour, 3 hours, and 6 hours for RS($7,8$), RS($11,11$), and RS($15,15$), respectively. POMCGS can converge to near-optimal solutions similar to SARSOP on RS($7,8$) and RS($11,11$).
On the largest domain, RS($15,15$), SARSOP cannot run successfully (the model being too large to generate and store), while POMCGS is able to obtain competitive performance compared with state-of-the-art online solvers.
\subsubsection{Light Dark:}
\citet{Sunberg_2017} propose a continuous-observation problem where the agent moves in a one-dimensional space, trying to reach a target region. 
The observation of the location is only accurate in a specific ``light'' region and is much more noisy in dark regions. 

%
%

We set a 1-hour timeout for offline solvers on the Light Dark problem. 
Under these conditions, POMCGS achieves a value of $3.74$, closely matching the performance of AdaOPS and outperforming other offline and online algorithms.
Due to the 1-dimensional nature of the observations in Light Dark, the K-means method effectively clusters continuous observations and constructs corresponding beliefs, contributing to POMCGS's strong performance. 
Further improvements may be achievable by increasing the value of $K$, provided that sufficient computational resources are available.


%
%
%
%

\subsubsection{Laser Tag:}
In Laser Tag \citep{NIPS2013_c2aee861}, a robot moves in a $7 \times 11$ grid map (with randomly placed obstacles) and tries to tag an escaping target.
In the beginning, the robot knows neither its own location nor the target location.
Moreover, the robot is equipped with a laser sensor that reads noisy distances in eight directions.
This creates a large number of possible eight-dimensional observations (around $1.5 \times 10^6$).
%

%
We use POMCGS together with observation clustering with a 3-hour timeout on Laser Tag due to its large observation space.
Interestingly, although the state space in Laser Tag is not extremely large, it turns out to be the most difficult problem for our offline method.
This may be due for instance
\begin{enumerate*}
\item to the 8-dimensional observation space possibly preventing from obtaining relevant clusters, and/or
\item to the merging of nodes, which may prevent necessary belief distinctions (while the agent needs to progressively refine its knowledge).
\end{enumerate*}
%
%
%
%
%
%
%
%
%
%
%
%
As a result, given a bounded running time, POMCGS is not able to compute a complete policy to tag the target, which leads to a poor performance.

\subsubsection{Bumper Roomba and Lidar Roomba:}
In the \textit{Roomba} problem \citep{roomba}, a cleaning robot is placed in a known room with an unknown initial position.
The robot needs to localize itself and reach a target region.
There are two sensing options for the robot depending on the problem: a Bumper sensor that can detect collisions, or a Lidar sensor that reads a noisy distance to the front.
%

%
%

We set a timeout of 3 hours for POMCGS in these two benchmark domains\footnote{\citet{NEURIPS2021_ef41d488} experimented with AdaOPS on Lidar Roomba and Bumper Roomba with finite-action sets and compared it with DESPOT.}.
Both POMCGS and POMCPOW can successfully localize and navigate the robot to the target, achieving similar performance in the Lidar Roomba problem.
%
%
%
In the Bumper Roomba problem, POMCGS converges to a positive value of 0.31, while we are unable to tune POMCPOW to have a similar performance.
During simulations, we observe that our offline policy is able to take correct decisions earlier than POMCPOW, and, more importantly, has less failing executions that navigate the robot to the wrong target.
%
%
POMCGS may benefit from never using belief estimates relying on very few particles, and from its FSC structure to update action values efficiently.

\subsection{A Closer Look at POMCGS's Behavior}

Given a fixed computation time, we observe that parameters ($\xi$, $K$, and $nb_{\text{particles}}$), which influence belief formation, have a strong impact on the final results.
Among these parameters, $\xi$ represents the belief merging threshold and directly affects the accuracy of the belief comparison process;
$K$ controls observation clustering, influencing the selection of samples for constructing new beliefs in continuous-observation POMDPs;
and $nb_{\text{particles}}$ defines the number of samples in a belief, impacting the accuracy of belief estimation.
To better understand POMCGS's behavior, this section provides an analysis of the influence of parameters $\xi$, $K$, and $nb_{\text{particles}}$.

\begin{figure}
  \centering
    \includegraphics[width=0.48\textwidth]{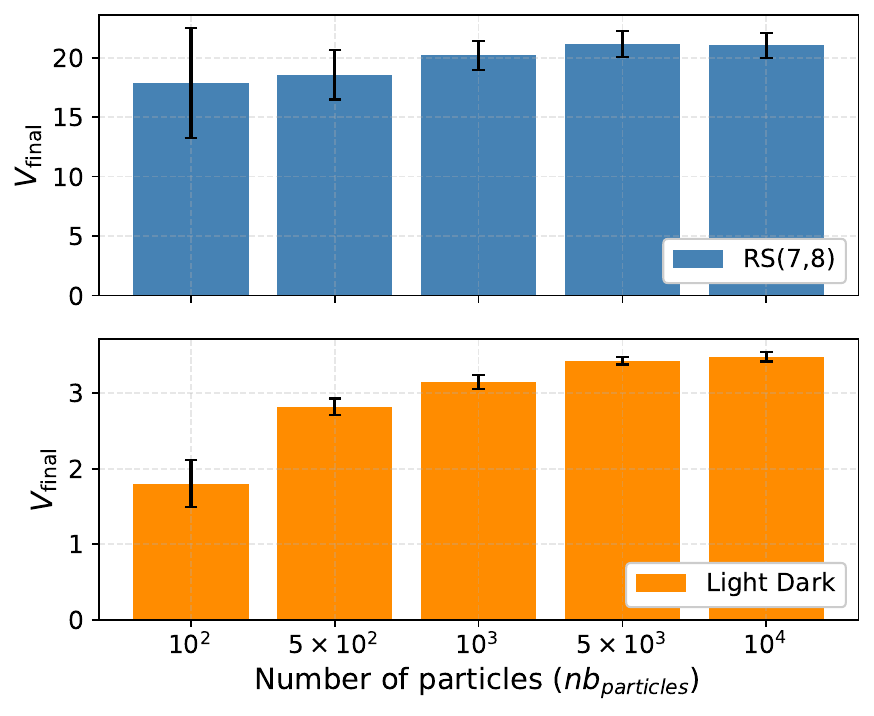}
  \caption{POMCGS's performances on  RS($7, 8$) and Light Dark problems with 5 different $nb_{\text{particles}}$ values.}
  \label{fig:nb_particle_test}
\end{figure}

We first investigate the influence of the number of particles (\(nb_{\text{particles}}\)) on the performance of POMCGS. We select two problems: RS\((7, 8)\) and Light Dark, fixing \(\xi = 0.1\) for both problems and setting \(K = 5\) for observation clustering in the Light Dark problem. All experiments are conducted with a timeout of 1 hour.
We evaluate five different values of \(nb_{\text{particles}}\). Our findings indicate that increasing the number of particles improves belief estimation accuracy. As shown in \Cref{fig:nb_particle_test}, we observe that a higher number of particles consistently leads to better performance. 
In smaller domains, such as RS\((7, 8)\), even a relatively small number of particles per node can yield competitive solutions. However, in more complex problems, a larger \(nb_{\text{particles}}\) is generally necessary to accurately represent beliefs. 
Based on these observations, we fix \(nb_{\text{particles}}\) at \(10^4\) for subsequent tests.

\begin{figure*}
\centering
\begin{subfigure}{.329\textwidth}
    \centering
    \includegraphics[width=.99\linewidth]{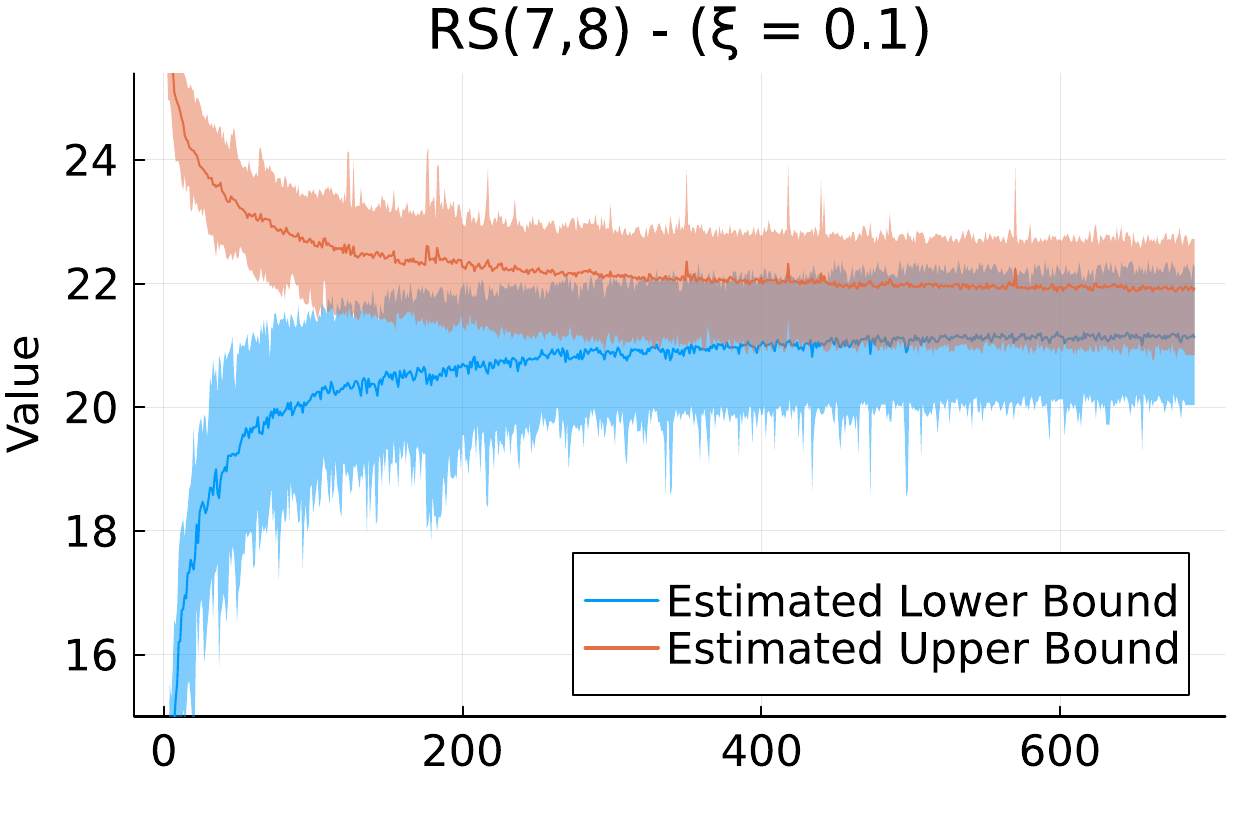}
    \label{SUBFIGURE LABEL 1}
\end{subfigure}
\begin{subfigure}{.329\textwidth}
    \centering
    \includegraphics[width=.99\linewidth]{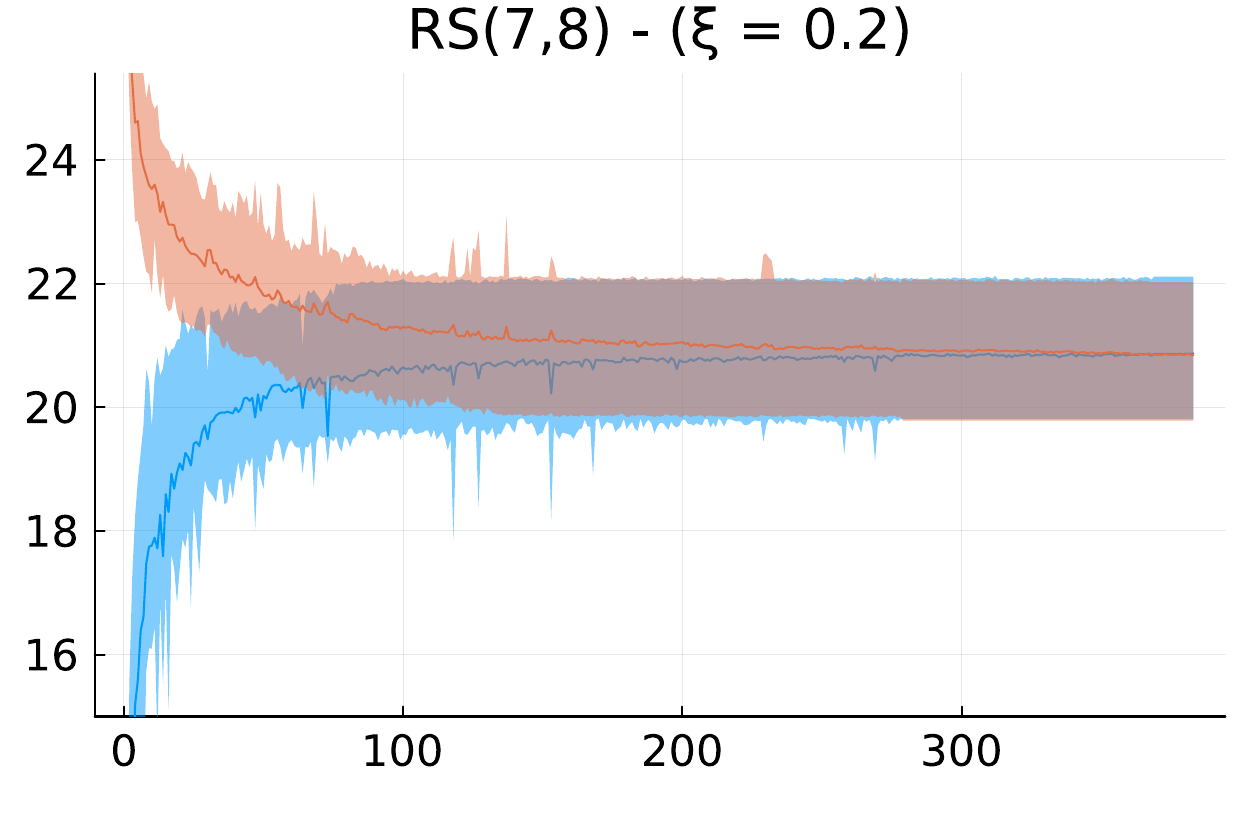}
    \label{SUBFIGURE LABEL 2}
\end{subfigure}
\begin{subfigure}{.329\textwidth}
    \centering
    \includegraphics[width=.99\linewidth]{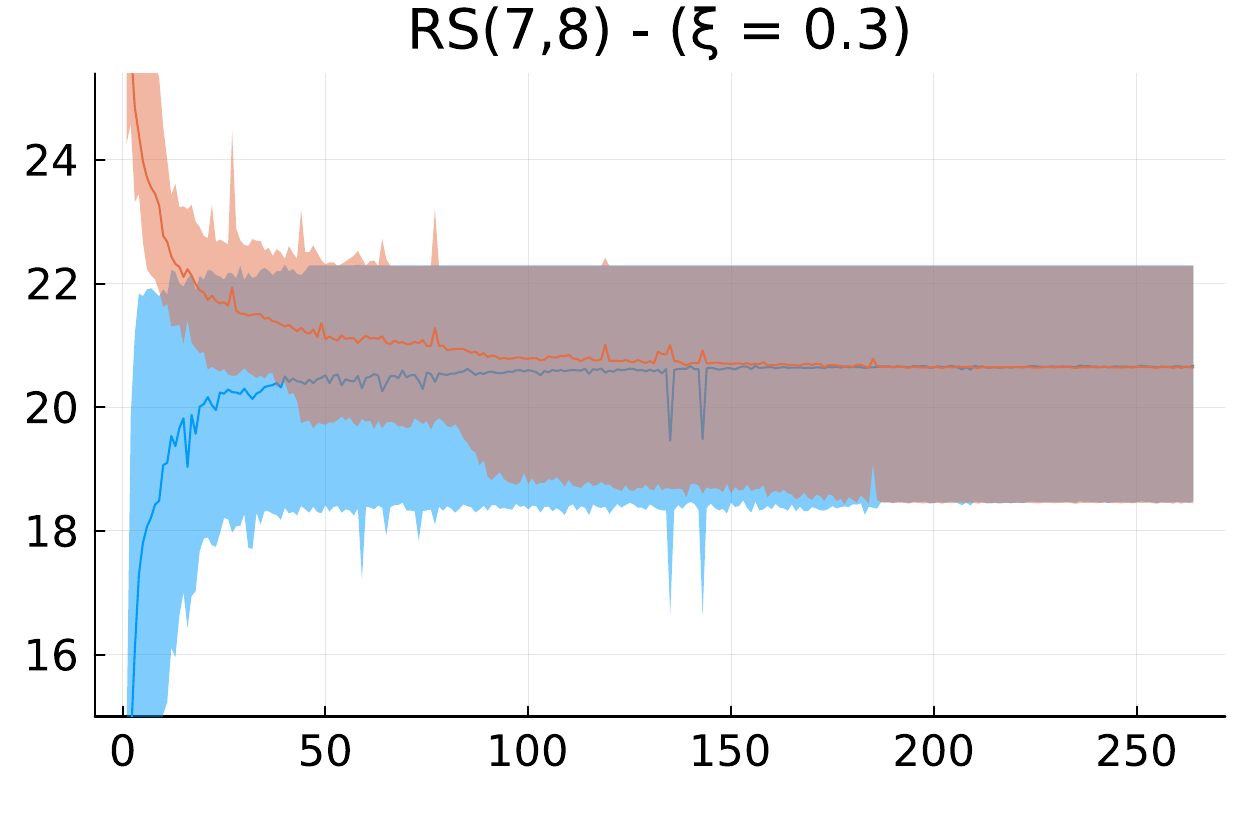}
    \label{SUBFIGURE LABEL 3}
\end{subfigure}

\begin{subfigure}{.329\textwidth}
    \centering
    \includegraphics[width=.98\linewidth]{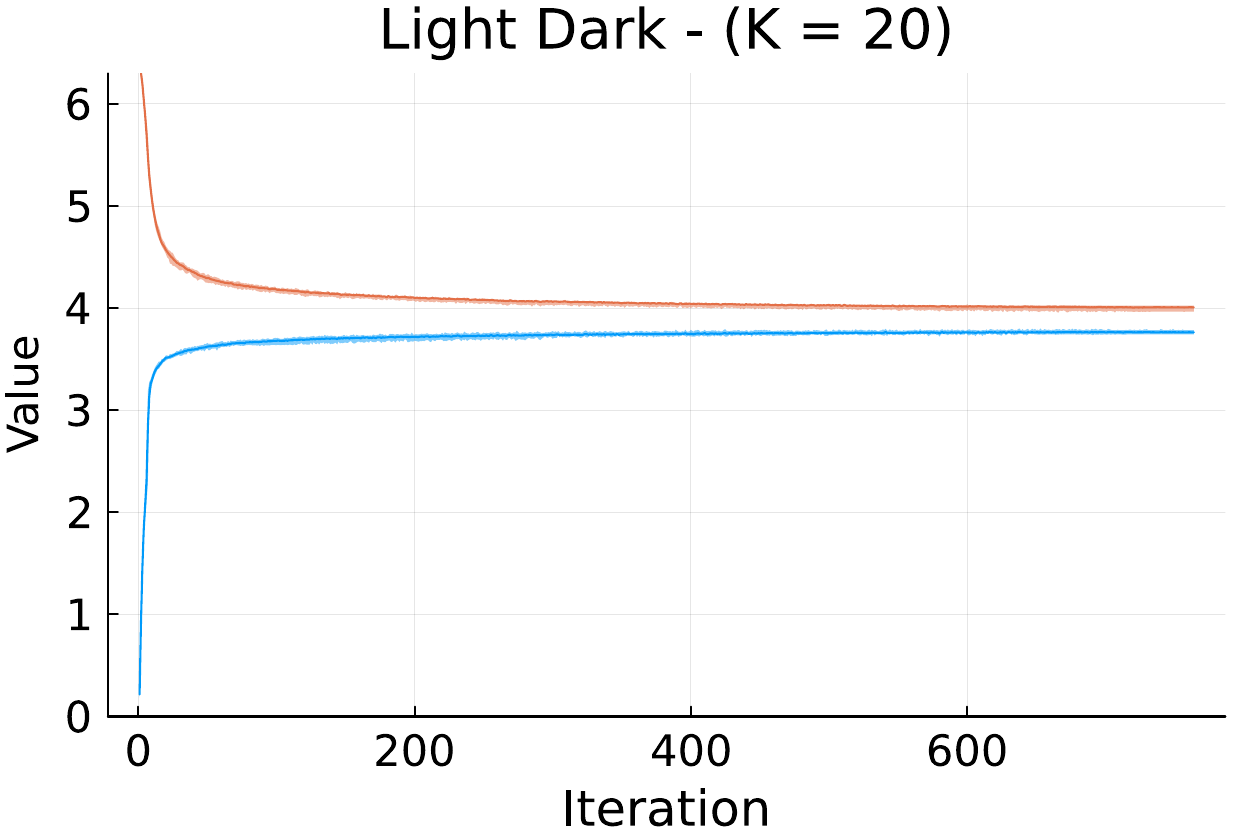}
    \label{SUBFIGURE LABEL 1}
\end{subfigure}
\begin{subfigure}{.329\textwidth}
    \centering
    \includegraphics[width=.98\linewidth]{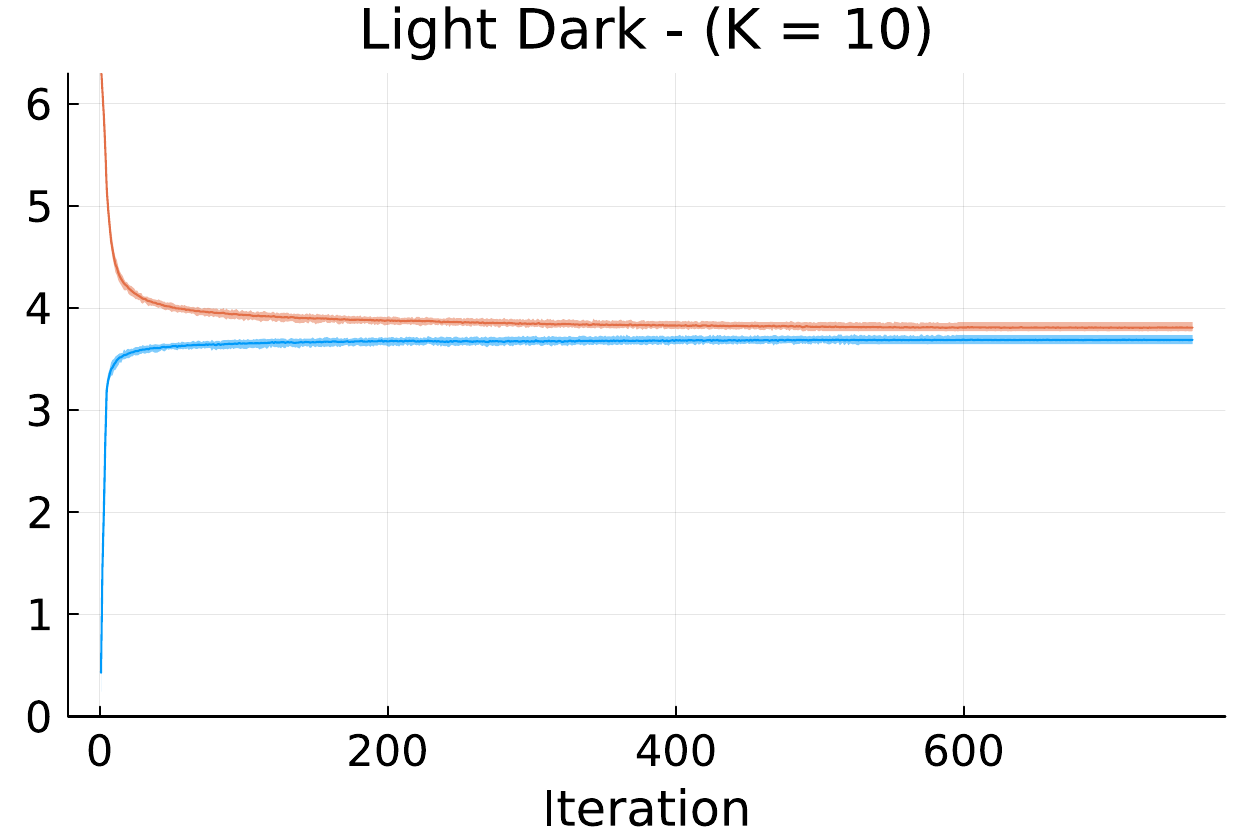}
    \label{SUBFIGURE LABEL 2}
\end{subfigure}
\begin{subfigure}{.329\textwidth}
    \centering
    \includegraphics[width=.98\linewidth]{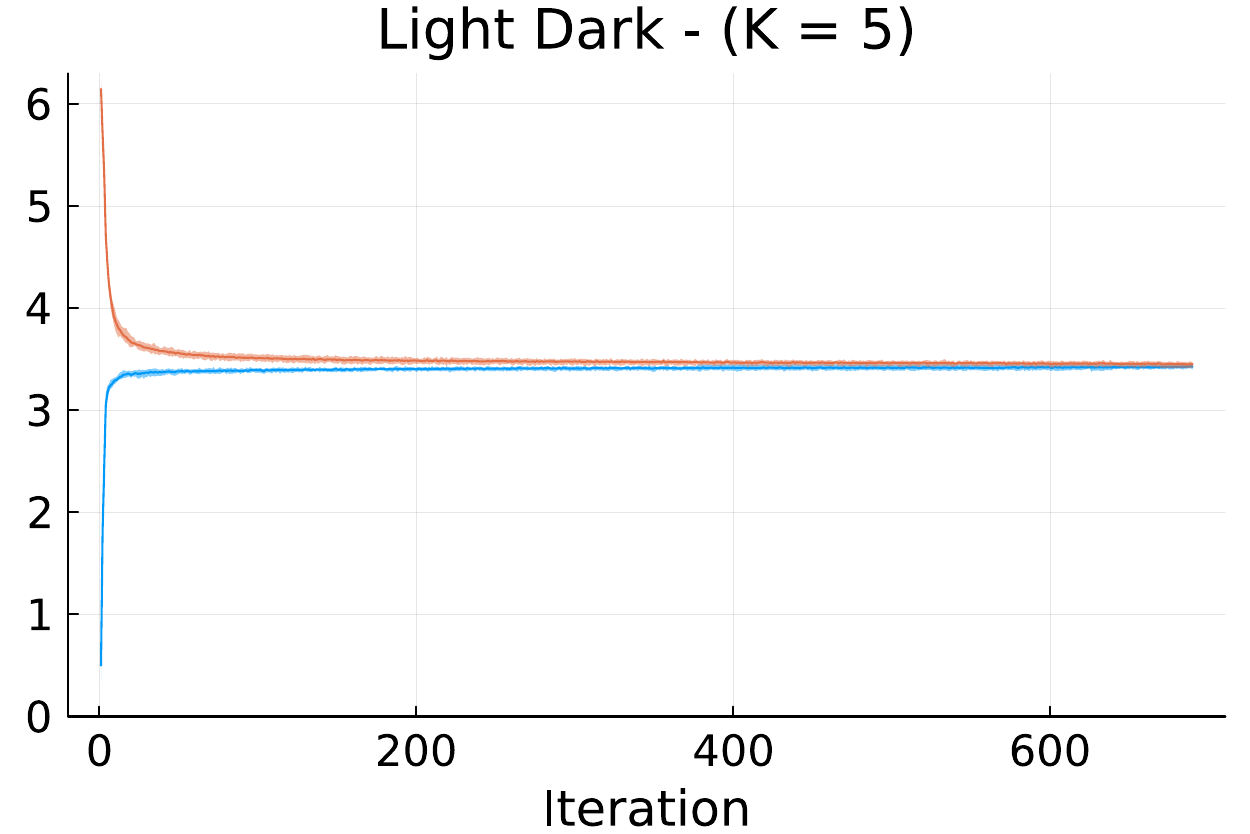}
    \label{SUBFIGURE LABEL 3}
\end{subfigure}
\caption{
Figures present the evolution of the POMCGS's upper and lower bounds at each iteration for the RS(7,8) and Light Dark with different parameters. 
}
\label{fig:results}
\end{figure*}

We then study the POMCGS's performance with three different belief merging thresholds on RS($7, 8$). As illustrated in \Cref{fig:results}, when $\xi$ increases from $0.1$ to $0.3$, fewer iterations are required to converge, but this comes at the cost of decreased value because the belief merging becomes less accurate.
Especially, POMCGS has the highest lower bound value when $\xi = 0.1$, but its lower bound and upper bound have not converged yet.
%
We next study the influence of the parameter $K$ for observation clustering. In Light Dark, we keep $\xi = 0.1$ and test three $K$ values ($5, 10, 20$). The POMCGS's behavior follows a similar trend where fewer observation clusters result in a lower convergence value and fewer iterations.

We observe that tuning $\xi$ and $K$ allows for a trade-off between producing high solution quality when $\xi$ is small and $K$ is large, which requires more planning time (iterations) to converge, and achieving faster convergence but lower solution quality when $\xi$ is large and $K$ is small.
How to tune $\xi$ and $K$ thus usually depends on the problem at hand with respect to the solution quality and computational resources.
In this article, we propose a standard parameter configuration with respect to the problem size as shown in \Cref{tab:problem_types}, which is also used to report POMCGS's performance in \Cref{table1}.
One may achieve better performance by tuning these parameters when computational resources are available.

\begin{table}[h]
    \centering
    \setlength{\tabcolsep}{4pt} 
    \resizebox{\columnwidth}{!}{
    \begin{tabular}{c c c c c}
        \toprule
        \textbf{Problem Type} & \textbf{Examples} & {$nb_{\text{particles}}$} & {$\xi$} & {$K$ (if needed)} \\
        \midrule
        Small-to-medium & \parbox{4cm}{\centering RS(7,8)\\ RS(11,11)\\ Light Dark\\ Bumper Roomba} & $5 \times 10^3$ & 0.1 & 10 \\
        \midrule
        Large & \parbox{4cm}{\centering RS(15,15)\\ Lidar Roomba\\ Laser Tag} & $10^4$ & 0.3 & 5 \\
        \bottomrule
    \end{tabular}
    }
    \caption{Parameter Settings for Different Problems}
    \label{tab:problem_types}
\end{table}



%

%

\section{Discussion}
\label{sec:discussion}

\paragraph{Contribution:}

POMCGS differs from previous state-of-the-art offline solvers, which perform backup operations on alpha-vectors or policy graphs, as it does not adhere to the value iteration scheme.
Instead, POMCGS directly searches and updates within the same policy graph. 
The core innovation of POMCGS is its ability to merge similar beliefs within a Monte Carlo search tree, producing offline policies represented as finite-state controllers (FSCs). 
This approach reduces redundant computations and facilitates efficient updates of node values within the FSC structure.
Our experiments demonstrate that POMCGS scales effectively across challenging benchmarks where prior offline solvers, such as SARSOP, cannot operate.
Furthermore, unlike previous offline methods, POMCGS achieves competitive performance with state-of-the-art online methods on most complex POMDPs, particularly in scenarios with small or low-dimensional continuous observation spaces.
The results also indicate that, despite the relatively coarse node merging, large state spaces do not pose significant challenges for POMCGS.

\paragraph{Limitation:}
While POMCGS enhances offline planning performance on large problems, we acknowledge that no single algorithm can address all possible scenarios in this field. 
One notable limitation of POMCGS, as well as other potential offline methods, is handling high-dimensional observation spaces (e.g., in the Laser Tag problem). 
Unlike online planners, which can employ observation progressive widening and reason about actual observations during execution, offline algorithms like POMCGS must discretize observations for both planning and execution. 
This constraint presents significant challenges for high-dimensional data, as effective observation discretization (e.g., using K-Means clustering) becomes increasingly difficult.

\paragraph{Convergence:}
\citet{DBLP:journals/jair/LimBKTS23} demonstrate that, when any online sampling-based solving algorithm for MDPs can be adapted to solving a POMDP through deriving an associated {\em Particle Belief MDP} (PB-MDP), and convergence guarantees of the algorithm (\ie, concentration bounds exploiting the Lipschitz continuity of the optimal value function) extend to POMDPs.
In this paper, \Cref{alg:UpdateFSC} can also be interpreted as relying on a PB-MDP, but as an offline algorithm, it must anticipate and discretize all possible continuous observations, employing a node merging process that results in an FSC rather than a tree or directed acyclic graph.
Proving convergence guarantees thus remains an open question.

\paragraph{Future Work:} 
We believe that offline planning in POMDPs with high-dimensional observations is a critical area deserving of dedicated research.
In this direction, we aim to improve the observation clustering process by incorporating dimensionality reduction techniques.
However, care must be taken regarding potential information loss, which could impair the algorithm's ability to distinguish between beliefs.
Another area of future work could explore avoiding state discretization by using more effective belief comparison metrics.

\section{Conclusion}
\label{sec:conclusion}

In the last decade, most POMDP planning methods have focused on solving large problems online. 
However, in extreme environments or under severe constraints, embedded systems often lack the resources needed for online computation.
At the same time, traditional offline solvers struggle to scale up for large problems or to provide effective solutions in continuous domains.
In this article, we present a novel offline POMDP algorithm, POMCGS, to address these challenging POMDPs and obtain compact policies that can be directly executed without further computations.
Relying on Monte-Carlo simulations, we relax the requirements of explicit models and successfully turn the search tree into a policy-graph (\ie, a finite-state controller).
In particular, we pave the way for a new approach, distinct from previous (point-based) value iteration methods, to compute complete policies in large, continuous domains.


\section*{Acknowledgments}

This work has been supported by the EPSRC Energy Programme under UKAEA/EPSRC Fusion Grant 2022/2027 No. EP/W006839/1 and by French National Research Agency (ANR) through the “Flying Coworker” Project under Grant 18-CE33-0001.

\bibliography{refs-short,refs}

\end{document}